\documentclass[letterpaper,10pt,conference]{ieeeconf}  

\IEEEoverridecommandlockouts                              

\usepackage{graphics} 
\usepackage{mathptmx} 
\usepackage{amsmath} 
\usepackage{amssymb}  
\usepackage{algpseudocode}
\usepackage{todonotes}
\usepackage{tikz}
\usetikzlibrary{positioning, arrows.meta}
\usetikzlibrary{calc}

\usepackage{amsthm}

\usepackage{dsfont}
\usepackage{algorithm}
\usepackage{xcolor}
\let\labelindent\undefined
\usepackage[inline]{enumitem}

\usepackage{subfig} 
\usepackage{array}  
\usepackage{booktabs}  
\usepackage{tikz}
\usepackage{pgf}
\usepackage{pgfplots}

\usetikzlibrary{tikzmark,decorations.pathreplacing,calligraphy,arrows,shadows,petri,automata,shapes,shadows,trees,fit, calc}
\tikzset{
	basic/.style  = {draw, text width=2cm, drop shadow, font=\sffamily, rectangle},
	root/.style   = {basic, rounded corners=2pt, thin, align=center,
		fill=green!30},
	level 2/.style = {basic, rounded corners=6pt, thin,align=center, fill=green!60,
		text width=8em},
	level 3/.style = {basic, thin, align=left, fill=pink!60, text width=8.5em}
}
\tikzset{myarrow/.style={-{Latex[length=2mm]}}}
\usepgfplotslibrary{groupplots}
\usetikzlibrary{shapes,arrows}
\usetikzlibrary{decorations.pathreplacing}
\usetikzlibrary{babel}
\usetikzlibrary{calc,arrows.meta,patterns,backgrounds}
\usetikzlibrary{positioning}
\usetikzlibrary{fit}

\tikzstyle{block}  =  [draw,  rectangle,  minimum  height=2em,  minimum  width=4em]
\tikzstyle{sum} = [draw, fill=blue!20, circle, node distance=1cm]
\tikzstyle{input} = [coordinate] \tikzstyle{output} = [coordinate]
\tikzstyle{pinstyle} = [pin edge={to-,thin,black}]
\usepackage[customcolors,norndcorners]{hf-tikz}
\pgfplotsset{compat=1.8}

\title{\LARGE \bf
Guided by Guardrails: Control Barrier Functions as Safety Instructors for Robotic Learning}

\author{Maeva Guerrier$^{1, 2}$, Karthik Soma$^{1, 2}$, Hassan Fouad$^{1}$ and Giovanni Beltrame$^{1,2 }$
\thanks{*This work was supported by CRIAQ.}
\thanks{$^{1}$The authors are with the Department of Computer Engineering and Software Engineering, Polytechnique Montreal, Montreal, Canada.
        {\tt\small maeva.guerrier@polymtl.ca}}%
\thanks{$^{2}$ Affiliated with Mila - Quebec AI Institute, Montreal, Canada.}%
}

\begin{document}

\maketitle
\thispagestyle{empty}
\pagestyle{empty}

\begin{abstract}

Safety stands as the primary obstacle preventing the widespread adoption of learning-based robotic systems in our daily lives. While reinforcement learning (RL) shows promise as an effective robot learning paradigm, conventional RL frameworks often model safety by using single scalar negative rewards with immediate episode termination, failing to capture the temporal consequences of unsafe actions (e.g., sustained collision damage).
In this work, we introduce a novel approach that simulates these temporal effects by applying continuous negative rewards without episode termination. Our experiments reveal that standard RL methods struggle with this model, as the accumulated negative values in unsafe zones create learning barriers. To address this challenge, we demonstrate how Control Barrier Functions (CBFs), with their proven safety guarantees, effectively help robots avoid catastrophic regions while enhancing learning outcomes.
We present three CBF-based approaches, each integrating traditional RL methods with Control Barrier Functions, guiding the agent to learn safe behavior. Our empirical analysis, conducted in both simulated environments and real-world settings using a four-wheel differential drive robot, explores the possibilities of employing these approaches for safe robotic learning.




\end{abstract}

\section{INTRODUCTION}

Artificial Intelligence (AI) powered robotic systems have applications across diverse domains including agriculture, healthcare, and space exploration. However, imagine a search and rescue robot that repeatedly collides with trees, becomes stranded, and fails its mission. This scenario highlights a critical question: How do we enable safety-based learning in autonomous systems?

Ensuring safety in learning methods remains an open challenge~\cite{bengio2024internationalscientificreportsafety, barez2025openproblemsmachineunlearning}, that safe-by-design AI aims to address. Learned policies enable generalizable behaviors, while safety prevents harmful outcomes in real-world. However, learning and safety might be in conflict in certain scenarios, as the pursuit of optimal performance may often lead to unsafe behaviors~\cite{cohen2024rldontiwouldnt}.  

\begin{figure}[hbt]
	\centering
	\begin{tikzpicture}[auto, node distance=2cm,>=latex',scale=1.2,transform shape]
		\input{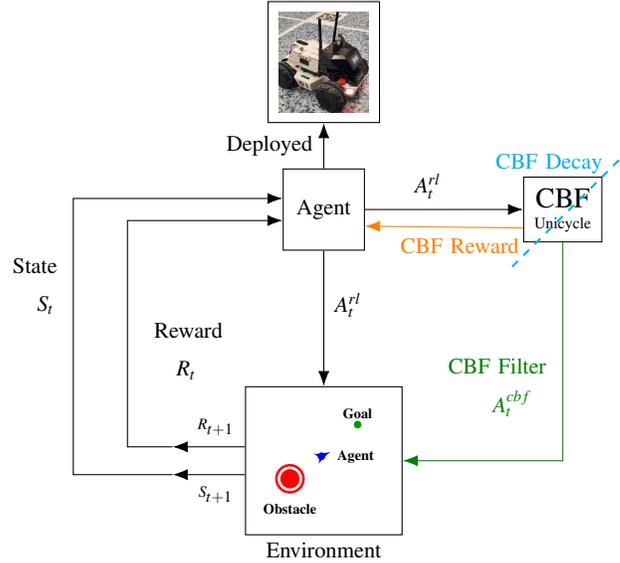}
	\end{tikzpicture}
	\caption{Overview of the three variations of safety guardrail for reinforcement learning: a filter (green) approach that intervenes when the agent is about to enter an unsafe region, a reward-based (orange) approach that incorporates CBF within the reward structure and a decay (blue) approach that progressively removes the effect of the CBF over time.}
	\label{fig:front_page}
\end{figure}

Among various learning paradigms, Reinforcement learning (RL) \cite{Sutton_RL_Intro} has emerged as one of the efficient methods that enables the learning of intelligent behaviors for robots \cite{kaufmann2023champion}. 
Traditional RL frameworks typically represent unsafe situations using negative reward signals followed by immediate episode termination \cite{kobayashi2023intentionallyunderestimatedvaluefunctionterminal, pham2018constrained}. Yet, this approach fails to accurately reflect real-world catastrophic scenarios and their consequences over time, as these artificial terminations don't mirror natural outcomes. The problem is compounded by RL's inherent exploration mechanism: without proper boundaries, agents can become trapped in catastrophic states, significantly impeding the learning process. These limitations underscore the critical importance of incorporating safety constraints within RL frameworks.

However, ensuring safety remains inherently difficult, and implementing safety constraints within the learning process is even more challenging. To address this challenge, Control Barrier Functions (CBFs~\cite{ames2019controlbarrierfunctionstheory}) have been widely applied in robotics to enforce safety. CBFs offer a theoretically grounded framework to enforce safety invariance~\cite{CBF_Introduced_By_Weiland, ames2019controlbarrierfunctionstheory}.
Nevertheless, synthesizing CBFs is complex as it requires complete knowledge of system dynamics, demanding extensive domain expertise. For practicality, CBFs that employ basic models to abstract robot dynamics are generally preferred.
Our objective is to integrate CBFs (see Figure \ref{fig:front_page}) as a safety guardrail into RL to develop safe behaviors. Specifically, we aim at learning both goal-reaching and safety-maintaining policies for robots. We adopt our definition from~\cite{bengio2024bayesianoraclepreventharm}: a safety guardrail evaluates the admissibility of actions in the current state, restricting actions that can lead to failures. 

Considering a goal-reaching and obstacle avoidance as the robot's task, the main contributions of this paper are: 
\begin{itemize}
    \item We design a per-step CBF that integrates seamlessly with the RL observation-action framework by intervening when unsafe actions are selected by the agent.
    We chose a unicycle model that enables the simplification of the CBF formulation and it also provides a representation model that easily generalizes to many robots, as the abstraction it provides is compatible for a wide range of robot dynamics. We demonstrate this by deploying the learned policy from the abstracted unicycle model trained in simulations onto a four-wheel differential drive robot.
    \item We propose three RL-CBF integration methods: (1) CBF Filter, a CBF-based action filter; (2) CBF Reward, which incorporates CBF into the reward function; and (3) CBF Decay, where the influence of CBF diminishes over time. We evaluate these approaches using multiple metrics (e.g., rewards, CBF activations) and analyze their sim2real transfer performance, providing insights into the learned behaviors of each variant.
\end{itemize}

\section{RELATED WORK}

\textbf{Safety in RL (SRL).} 
The exploration-exploitation dilemma of RL \cite{efroni2020explorationexploitation} concerns balancing the discovery of new knowledge against exploiting existing information to maximize performance, but exploration can lead to hazardous actions. Early works \cite{rl_safe_survey_Garcia_2015} emphasized the need for safety constraints during exploration. Recent advances in SRL focuses on Constrained Markov Decision Processes (CMDPs) to enable safe behaviors \cite{rl_safe_survey_2024}. While most CMDP approaches prioritize intermediate costs \cite{achiam2017constrained,Reward_Constrained_Policy_Optimization}, they often fail to account for catastrophic events, which better reflect real-life scenarios.
Some approaches incorporate safety directly into the optimization objective~\cite{achiam2017constrained,Reward_Constrained_Policy_Optimization, miryoosefi2021simplerewardfreeapproachconstrained}, while others derive strategies from game theory to address safety concerns~\cite{Reinforcement_Learning_Convex_Constraints}. Additionally, certain methods leverage external knowledge to enforce safety constraints~\cite{bengio2024bayesianoraclepreventharm, Sim_to_Lab_to_Real_Safe_Reinforcement_Learning_with_Shielding_and_Generalization_Guarantees}.
Recent years have witnessed the emergence of using safety layer \cite{Safe_Reinforcement_Learning_via_Shielding, robustModelPredCtrlShieldforRL} in RL methods. 



\textbf{Linking CBFs and Safety Guardrails.} There is an extensive literature on CBFs for safety critical systems \cite{ames2019controlbarrierfunctionstheory, rl_guerrier2024learningcontrolbarrierfunctions, SafeLearningCBFandCLFSurvey}. 
Subsequent research~\cite{amesCBFBasedQuadraticSafetyCritSyst, ames2019controlbarrierfunctionstheory} demonstrated how CBFs can be integrated with control policies to guarantee safety while optimizing performance. Taylor et al.~\cite{taylor2019learningsafetycriticalcontrolcontrol} extended CBFs to high-dimensional systems, making them applicable to a wider range of real-world problems, including robotics and autonomous agents. The concept of safety guardrails~\cite{dong2024buildingguardrailslargelanguage} aligns closely with the principles of CBFs, as both aim to enforce safety constraints dynamically.


\textbf{CBFs in RL.}
Recent surveys~\cite{rl_safe_survey_2024, rl_guerrier2024learningcontrolbarrierfunctions} examine integrating CBFs in the RL framework for providing safety guarantees. 
Waberish et al.\cite{wabersich2021predictivesafetyfilterlearningbased} employ neural network-based CBF controllers instead of solving quadratic programs at each step. However, neural CBFs introduce non-differentiability issues that can compromise safety guarantees. Alternative approaches use Gaussian Processes to handle disturbances and model uncertainty \cite{End_to_End_Safe_Reinforcement_Learning_through_Barrier_Functions_for_Safety_Critical_Continuous_Control_Tasks, emam2022safereinforcementlearningusingRCBF}.
Emam et al.~\cite{emam2022safereinforcementlearningusingRCBF} integrate CBFs directly into the value function for off-policy actor-critic methods. Their approach uses a conservative policy to limit exploration to safe regions, but ties the agent to specific dynamics models and requires small time steps, limiting real-world applicability. Our work addresses these limitations by introducing a CBF-based safety guardrail with a priority variation parameter, enabling larger time steps and proposing an abstract model that decouples CBF from specific system dynamics, enhancing applicability to complex robotic systems.

\section{PRELIMINARIES}


\subsection{Reinforcement Learning}

RL consists of a Markov Decision Process represented by the tuple ($S, A, R, P$), where $S$ is the sate space, $A$ the action space, $R$ the reward function, $P$ the state-transition probability, where an agent learns to solve a task depicted through rewards $r \in \mathds{R}$. The agent at any given state $s_t \in S$, samples an action from its policy $\pi(a|s_t)$ then transitions to a next state $s_{t+1}$ based on the transition probability function $P$ and receives a reward $r_t$ from the environment.
The agent aims to find the optimal policy $\pi^*$ that maximizes the cumulative discounted reward: named return $G_t = \sum_{t=0}^{\infty}\gamma^tr_{t}$  ($\gamma \in [0, 1)$ is the discount factor). Value function of a policy $\pi$ ($V^{\pi}(s) = E_{\pi} [ \sum^{\infty}_{t=0} \gamma^t (R(s_t, a_t, s_{t+1}) | s_0 = s]$) is generally used to denote the expected return from a state $s_t$. 
Action space $A$ and $S$ can be discrete or continuous. In this paper, we consider the Soft Actor-Critic (SAC) \cite{sac} off-policy RL algorithm \cite{sac}, which has two neural networks; Actor (the policy $\pi$) and Critic (the value function network).  

\subsection{Control Barrier Function}

Within a Control Barrier Functions (CBF) framework, the concept of safety is linked to a function \( h(x) \) such that \( h(x) \) is a continuously differentiable function where \( x \)\footnote{System's state $x$ can be viewed as $s_t$ in the RL framework.} is the system's state. 
The function \( h(x) \) can be applied to define the safe set such that, the system is considered safe if \( h(x) \geq 0 \). Usually, CBF deal with control affine systems:

\begin{equation}
\dot{x} = f(x) + g(x)u
\end{equation}

where \( u\) is the action passed to the system, \( f(x)\) is the system dynamics, \( g(x)\) is the input dynamics.

The goal of using a CBF is to identify an input \( u \) such that: 
\begin{equation}
\underbrace{L_fh(x) + L_gh(x)u}_{\dot{h}(x)}\geq -\alpha(h(x)) 
\end{equation}

where \( L_f h(x) = \nabla h(x) . f(x) \), and \( L_g h(x) = \nabla h(x) . g(x) \) are the Lie derivative of $h(x)$ with respect to system and input dynamics respectively. 

A common way to enforce safety on the output of a RL policy $\pi$ is to solve the following quadratic program:
\begin{equation}
	\begin{aligned}u^*=
		& \underset{u\in U}{\text{minimize}}
		& & \|u - k\|^2 \\
		& \quad\text{s.t.}
		& & L_f h(x) + L_g h(x) \, u \geq -\alpha(h(x))
	\end{aligned}
	\label{eqn:QP}
\end{equation}

where \( u^* \) is a safe action and \( k \sim \pi \) is the RL policy action.

\subsection{Obstacle Avoidance Using CBFs for RL}
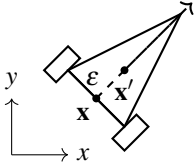
\begin{figure}[h]
    \centering
    \begin{tikzpicture}[scale=0.75]
    \coordinate (A) at (1,1);
    \coordinate (B) at (2,0);
    \coordinate (Mid) at (1.5,0.5);
    \coordinate (D) at (2.,1.);
    \coordinate (C) at (3.,2.);
    \coordinate (E) at (3.2,2.2);
    \pgfmathsetmacro{\theta}{45};
    \def\theta{45};
    \pgfmathsetmacro{\width}{0.6} ;  
    \pgfmathsetmacro{\height}{0.3} ; 
    \pgfmathsetmacro{\angle}{45} ; 
    \pgfmathsetmacro{\dx}{(\height/2)*cos(\angle)};  
    \pgfmathsetmacro{\dy}{(\height/2)*sin(\angle)} ; 
    \coordinate (RectOrigin) at ($(A) + (-\dx,\dy)$);
    \coordinate (RectOrigin2) at ($(B) + (\dx,-\dy)$);
    \draw[shift={(RectOrigin)},thick, rotate around={\angle:(0,0)}] 
        (-\width/2, -\height/2) rectangle (\width/2, \height/2);

    \draw[shift={(RectOrigin2)},thick, rotate around={\angle:(0,0)}] 
        (-\width/2, -\height/2) rectangle (\width/2, \height/2);
    
    \draw[thick] (A) -- (B);
    \draw[thick,dashed] (Mid) -- (C);
    \draw[->,thick,dashed] (D) -- (E);
    \draw[thick] (A) -- (B) -- (C) -- cycle;
    
    \fill[black] (Mid) circle (2pt);
    \node[below,xshift=-5pt] at (Mid) {$\mathbf{x}$};
    \fill[black] (D) circle (2pt);
    \node[below,xshift=0pt] at (D) {$\mathbf{x}^\prime$};
    \node[below,xshift=-12pt,yshift=3pt] at (D) {$\epsilon$};
    
    \draw[->] (0.,-0.5) -- (1.,-0.5) node[right] {$x$};
    \draw[->] (0.,-0.5) -- (0.,0.5) node[above] {$y$};

\end{tikzpicture}
    \caption{Schematic of unicycle model. Point $\mathbf{x}^\prime$ is shifted $\epsilon$ units along robot axis.}
    \label{fig:unicycle}
\end{figure}

We assume that the positions of obstacles are known, and we focus primarily on a unicycle robot setup, shown in Figure~\ref{fig:unicycle}, and modeled as \eqref{eqn:unicycle}.
\begin{equation}
        \dot{x} = v\cos\theta;\
        \dot{y} = v\sin\theta;\ \dot{\theta} = \omega
    \label{eqn:unicycle}
\end{equation}
where $\mathbf{x}=(x,y)\in\mathbb{R}^2$ is the robot's coordinates, and $v\in\mathcal{V}\subseteq\mathbb{R}$ is robot's linear velocity, and $\omega\in\Omega\subseteq\mathbb{R}$ is its angular velocity.

In this paper, we are particularly interested in incorporating CBF in the RL framework. Specifically in this setup we have a fixed linear velocity $v_{des}$ and the RL agent can control the angular velocity $\omega_{\pi}$. \\

A common CBF candidate used for obstacle avoidance is 
\begin{equation}
    h = \|\mathbf{x} - \mathbf{x}_0\|^2 - \delta^2
\end{equation}
where $\mathbf{x}_0$ is the position of the obstacle's center, and $\delta$ is a desired safe distance. However, using this candidate for \eqref{eqn:unicycle} makes it higher relative degree CBF ($h$ needs to be differentated twice till $\omega$ shows up, which we want to be our primary method of avoidance). 

We instead opt for choosing $\mathbf{x}^\prime$, a point shifted a small distance $\epsilon$ along the robots axis as showin in Figure~\ref{fig:unicycle}, for our CBF candidate which becomes 
\begin{equation}
    h = \|\mathbf{x}^\prime-\mathbf{x}_0\|^2-(\delta+\epsilon)^2
    \label{eqn:oa_cbf}
\end{equation}
Note that the safe distance required is augmented by $\epsilon$ to ensure the actual robot's center at $\mathbf{x}$ does not collide with the obstacle. The derivative of \eqref{eqn:oa_cbf} is 
\begin{equation}
    \dot{h} = (\mathbf{x}^\prime-\mathbf{x}_0)^T\dot{\mathbf{x}^\prime}
    \label{eqn:h_dot}
\end{equation}
and we note that the relation between $\mathbf{x}^\prime$ and $\mathbf{x}$ is 
\begin{equation}
    \begin{split}
        \mathbf{x}^\prime &= \mathbf{x} + R(\theta)\begin{bmatrix}\epsilon\\0\end{bmatrix}\\\Rightarrow \dot{\mathbf{x}^\prime}&=\dot{\mathbf{x}}+SR(\theta)\begin{bmatrix}\epsilon\\0\end{bmatrix}\omega
    \end{split}
    \label{eqn:x_dot}
\end{equation}
where $R(\theta)$ is a rotation matrix from local axes to global axes, and $S = \begin{bmatrix}
    0&-1\\1&0
\end{bmatrix}$ is a skew symmetric matrix. 
When we plug \eqref{eqn:x_dot} in \eqref{eqn:h_dot} we notice that \begin{enumerate*}
    \item $\dot{h}$ can be influenced right away by two control actions (a robot can slow down or rotate around an obstacle to avoid it) and 
    \item $h$ is relative degree 1 (we differentiate once to find all control inputs)
\end{enumerate*}
We then modify \eqref{eqn:QP} in the following manner
\begin{equation}
	\begin{aligned}u^*=
		& \underset{v\in \mathcal{V}, \omega\in\Omega}{\text{minimize}}
		& & \kappa(v-v_{des})^2+(\omega-\omega_\pi)^2 \\
		& \quad\text{s.t.}
		& & R(\theta)\begin{bmatrix}
		    1&0\\0&\epsilon
		\end{bmatrix}\begin{bmatrix}
		    v\\u
		\end{bmatrix} \geq -\alpha(h)
	\end{aligned}
	\label{eqn:QP2}
\end{equation}
where $\kappa>0$ is a quantity that we use to set the priority to use linear velocity for avoiding collision. When $\kappa$ increases, more emphasis will be put on $\omega$ for obstacle avoidance.

Controlling both linear and angular velocities in \eqref{eqn:QP2} has the advantage of making the robot capable of avoiding obstacles, even at limited control actions. 
This is due to the fact that a robot will be able to slow down, thus allowing it to achieve any desired radius of curvature.

\section{METHODOLOGY}
We use SAC as the base and build upon it for our three RL-CBF integration methods.

\subsection{CBF Filter} 
We integrate SAC with CBF, which acts as a guardrail. The CBF overrides the agent’s actions in a minimally invasive manner when the agent approaches an obstacle within a safe distance $\delta$. \\ \\
\begin{equation}
a = \begin{cases} 
a_{\text{cbf}} & \text{if } h(x) \leq 0 \quad \text{(safety guardrail activated)}, \\
a_{\text{rl}} & \text{otherwise},
\end{cases}
\end{equation}

This setup neither modifies the reward design nor introduces additional training losses. \\
 
\subsection{CBF Reward} 
In this approach, SAC is combined with a reward function that penalizes the agent for its deviation from the CBF-proposed safe action (only effective within the safe distance $\delta$). Additionally, the approach penalizes the agent for velocity deviations if the CBF estimates a velocity correction when the agent is within the safe distance.

\begin{equation}
    r = \begin{cases} 
        |v_{\text{cbf}} - v_{\text{des}}| + |a_{\text{cbf}} - a_{\text{rl}}| & \text{if h(x) $\textless$ 0} \\
        0, & \text{otherwise}
    \end{cases}
\end{equation}

It is to be noted that unlike CBF Filter, CBF Reward does not override RL’s actions.
   
\subsection{CBF Decay}
We follow a curriculum learning-like approach, where the action applied at any training time step is a weighted combination of CBF's proposed action ($a_{cbf}$) and RL's policy ($a_{rl}$) (only activated within the safe distance $\delta$). Over training, the influence of CBF decays ($\beta = \beta - 1/{T}$, $\beta$ initialized at 0) and only RL's actions are applied by the end of training.     
\begin{equation}
    a \leftarrow \beta \times a_{\text cbf} + (1 - \beta ) \times a_{\text rl}
\end{equation}


\section{ENVIRONMENT}

\subsection{Simulation Setup}

We define the goal-reaching and obstacle-avoidance task (see Table \ref{tab:env_param}) as an RL gymnasium \cite{gymnasium} environment with randomized positions for starting point, goal, and one obstacle, at every episode reset. The agent dynamics is abstracted as an unicycle model. This is a sparse reward setting, the agent receives a +10 reward upon reaching the goal and a -10 reward upon colliding with the obstacle, but this does not terminate the episode. Additionally, if the agent crosses the boundary of the arena, the episode is not truncated. The episode truncates after a maximum number of steps and terminates only if the agent reaches the goal. The observation space is composed of the relative vectors of the goal and obstacle with respect to the agent's position in the agent's frame. 
The action and state space are continuous. The RL agent controls only the unicycle’s angular velocity ($\omega$), while the linear velocity ($v$) might be controlled by the CBF, if the agent is within the safe distance, otherwise; it is set to a fixed value. The CBF is activated only 
when the agent is within the safe distance $\delta$ from the obstacle.

\begin{table}[h]
    \centering
    \renewcommand{\arraystretch}{1.2} 
    \caption{Environment parameters}
    \label{tab:env_param}
    \begin{tabular}{@{} l l @{}}
        \toprule
        \textbf{Parameter} & \textbf{Value} \\
        \midrule
        Arena size & 1.5 m x 1.5 m\\
        Obstacle radius & 0.2 m\\
        Goal radius & 0.05 m\\
        Agent radius & 0.2 m\\
        Agent velocity ($v_{des}$) & 0.2 m/s\\    
        Action space & [-0.7, 0.7] rad/s\\
        Simulation time step & 0.1 ($\Delta t$)\\
        Maximum steps & 1000\\
        Total training steps (T) & 5e6\\
        Priority $\kappa$ & 500\\
        Safe distance $\delta$ & 0.45 m\\
        \bottomrule
    \end{tabular}
\end{table}

\subsection{Real World Setup}

We deploy a policy trained on a simplified unicycle model to a more complex four-wheel differential drive robot (Limo Agilex shown in Figure \ref{fig:front_page}). The pose estimates required for computing the observations of the agent are given by a motion capture system (OptiTrack Cameras). The agent is deployed in an arena (1.5m x 1.5m) and the episode is truncated when it goes out of the bounds of the real arena. 

\section{EXPERIMENTS}

\begin{figure*}[thpb]
    \centering
    \subfloat{%
        \includegraphics[width=.231\textwidth]{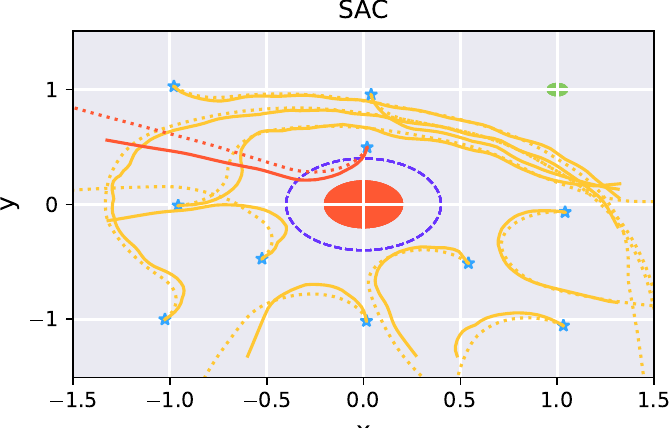}%
    }
    \subfloat{%
        \includegraphics[width=.222\textwidth]{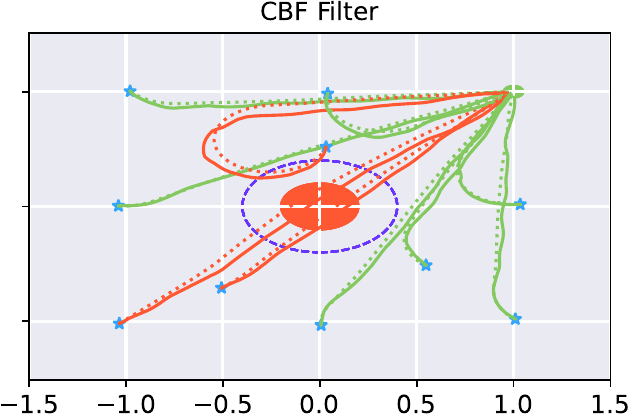}%
    }
    \subfloat{%
        \includegraphics[width=.222\textwidth]{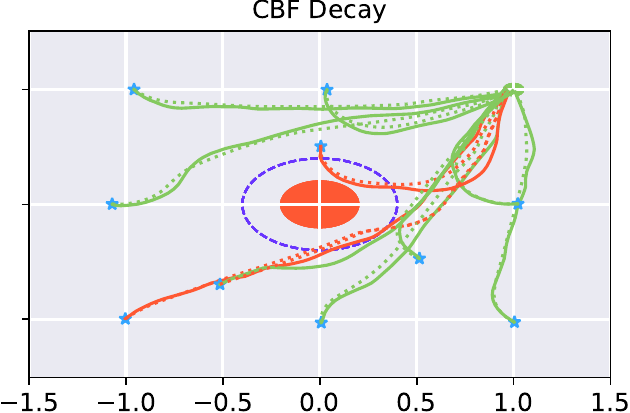}%
    }
    \subfloat{%
        \includegraphics[width=.222\textwidth]{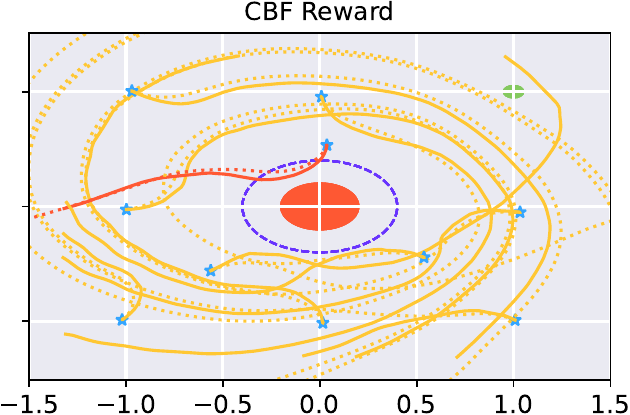}%
    }\\
    \subfloat{%
        \includegraphics[width=.75\textwidth]{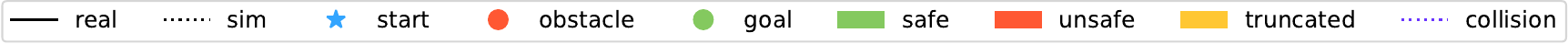}%
    }
    \caption{Simulated (dotted lines) and real (solid lines) trajectories w/o CBF for a four-wheel differential drive robot. The goal is the green circle, obstacle is a red circle with collision region (dot. purple circle). Trajectories can hit the obstacle (red) or not (green). Real trajectories are stopped if outside (orange) the arena; simulated ones stop after maximum steps is achieved (orange).}
    
    \label{fig:traj}
\end{figure*}

\begin{figure*}[thpb!]
    \centering
        \includegraphics[width=.26\textwidth]{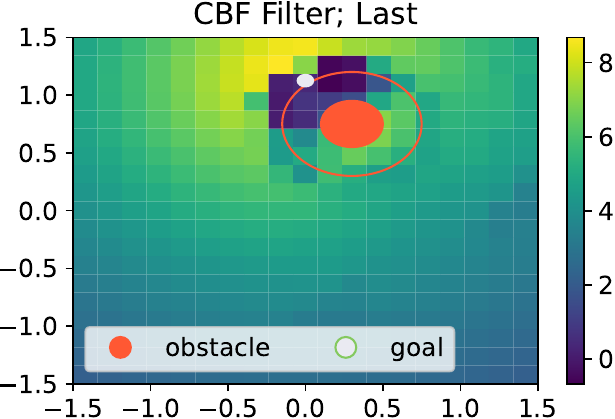}%
       \quad \includegraphics[width=.26\textwidth]{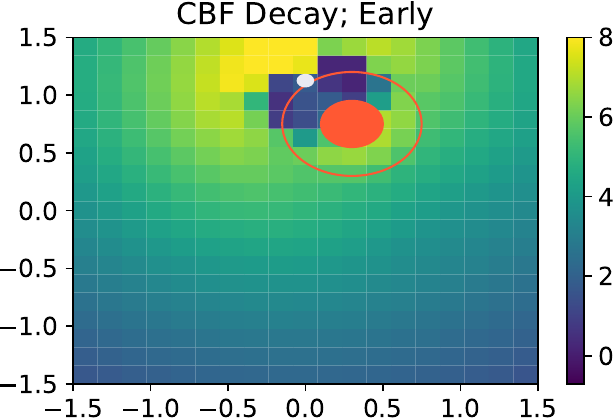}%
      \quad  \includegraphics[width=.28\textwidth]{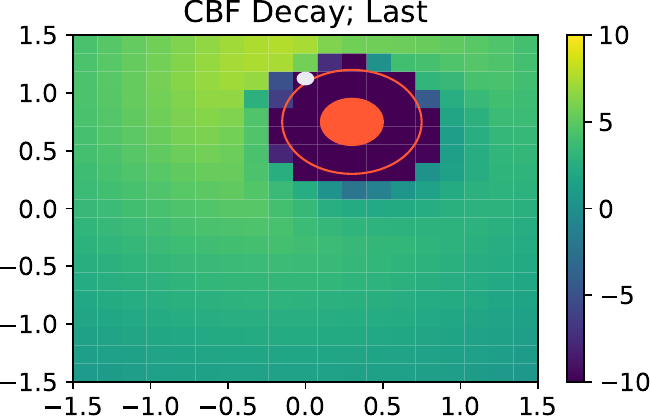}%
    \caption{Value of the critic for trained policies at early learning stage and the end of training. The agent is initialized at centroid of each bin with an orientation toward the obstacle.}
    \label{fig:value_func}
\end{figure*}

We evaluate the proposed three RL-CBF methods across 10 random seeds using various metrics, including the average reward per episode (Reward) and the percentage of CBF activations (Activations). The activations are measured as the percentage of times the agent enters the safe distance in an episode (activating the CBF, only with CBF filter and CBF decay). The policy is evaluated over 15 episodes at training steps, spaced at intervals of 100k steps (see Figure \ref{fig:rew_ep_length}). The average reward is measured in two scenarios: one with the CBF filter activated (\textit{w/ CBF}) and another without the CBF filter (\textit{w/o CBF}). Additionally, we calculate the critic's value function for observations corresponding to the centroid coordinates of discretized bins in the arena, with an orientation facing the obstacle. These values are averaged across all policies (10 seeds) at a given training step and visualized as a heatmap (see Figure \ref{fig:value_func}). For the sim2real deployment analysis, we visualize 10 real-world trajectories by deploying one of the learned policies (from the 10 seeds) on a four-wheel differential-drive robot, starting from various positions with an initial orientation facing the obstacle. We also collect corresponding simulation trajectories for the same starting points and visualize them in Figure \ref{fig:traj}.  


\section{RESULTS}
It is to be noted that if the CBF interventions are applied to the RL's action, it provides strong safety guarantees whether it is enabled during training in simulation or deployment in real world. However, in this section we are also interested in understanding the effects of removing the CBF during deployment and its impact on the learned policy. These are the insights we derive from our results:\\

\begin{figure*}[thpb!]
    \centering
        \includegraphics[width=.32\textwidth]{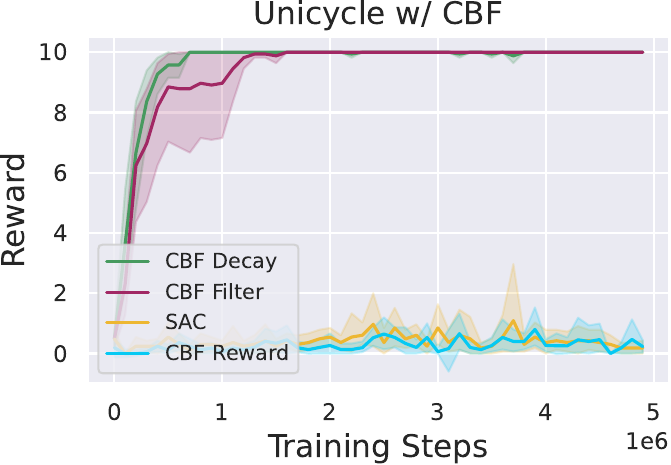}%
    \quad    \includegraphics[width=.3\textwidth]{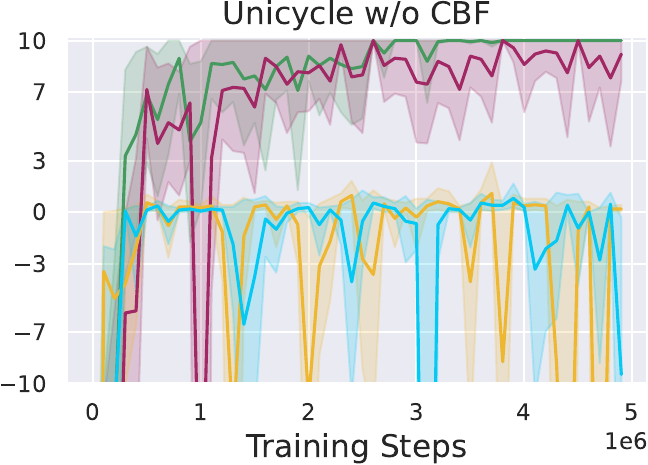}%
     \quad   \includegraphics[width=.32\textwidth]{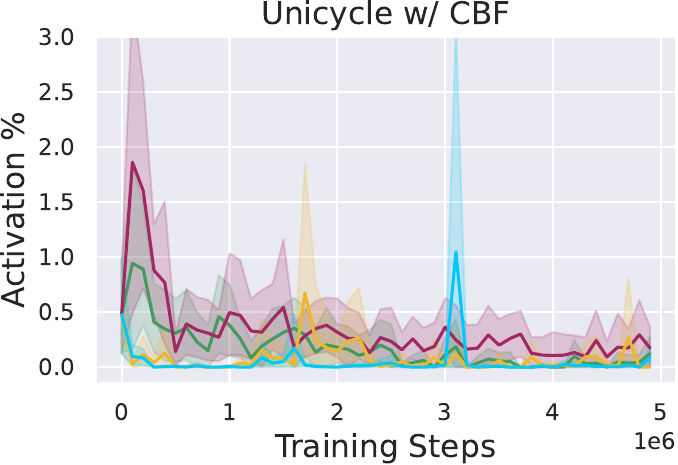}%
    \caption{Performance of the model with training: average reward, and average \% CBF activations per episode.}
    \label{fig:rew_ep_length}
\end{figure*}

\textbf{Catastrophic Regions Hinder Learning For SAC.} We observe from Figure \ref{fig:rew_ep_length} (both w/ \& w/o CBF) that SAC achieves near-zero returns, failing to learn how to reach the goal. 
Even with CBF protection enabled after training, SAC fails to reach the goal.
When we remove the CBF (\textit{w/o CBF}), we observe that SAC yields negative returns, indicating that the agent has not learned to avoid the obstacle always despite being able to explore the entire state space. Further, we observe in Figure \ref{fig:traj} (\textit{CBF Filter}) that SAC was deflected from the obstacle most times, but never reached the goal.

\textbf{CBF Filter Facilitates learning Goal-Reaching task but Fails to learn Obstacle Avoidance.} In contrast to SAC, the CBF Filter method learns to reach the goal, achieving the highest reward around $1\cdot 10^6$ training steps, as shown in Figure \ref{fig:rew_ep_length} (\textit{w/ CBF}). Both sim. and real. trajectories in Figure \ref{fig:traj} (\textit{CBF Filter}) confirm this, demonstrating the agent’s ability to reach the goal from any initial position.  While the CBF Filter enables the agent to reach the goal, when it is disabled, it does not achieve the highest reward because it collides with the obstacle, as reflected in Figure \ref{fig:rew_ep_length} (\textit{w/o CBF}). This behavior, where the agent becomes unaware of obstacles, is also evident in the value function visualization in Figure \ref{fig:value_func}, which shows high positive values inside the obstacle when the agent faces the goal. Further evidence of this phenomenon can be seen in both the simulation and real-world trajectories in Figure \ref{fig:traj} (\textit{CBF Filter}), where the agent moves through the obstacle. 

\textbf{CBF Reward: Ineffective for Goal-Reaching and Obstacle Avoidance.} CBF interventions do not teach the agent to recognize the dual components of the task: reaching the goal while avoiding obstacles. To address this, we introduce an additional reward that penalizes the agent if it does not imitate the CBF behavior near unsafe regions. Surprisingly, we can see in Figure \ref{fig:rew_ep_length} (\textit{w/ CBF}) that CBF Reward does not learn the task, as the rewards are around zero throughout the training. When we remove the CBF, in Figure \ref{fig:rew_ep_length} (\textit{w/o CBF}), we observe negative return, indicating that the agent might be colliding with obstacles. This clearly reveals that the agent has neither learned to reach the goal nor to avoid the obstacle. We observe this behavior in Figure \ref{fig:traj} (\textit{CBF Reward}), where the agent behaves randomly. In fact, this shows that relying solely on reward signals without the CBF's interventions hinders learning.

\textbf{CBF Decay: Demonstrate Goal-Reaching, Obstacle Avoidance or Recovery.} We see in Figure \ref{fig:rew_ep_length} (\textit{w/ CBF}) that CBF Decay gradually converges to the optimal return, indicating that the agent learned to reach the goal (faster than CBF Filter). This behavior is preserved when removing the CBF in \ref{fig:rew_ep_length} (\textit{w/o CBF}), where toward the end of training the agent is receiving the maximum possible reward. This demonstrates that the agent learned how to reach the goal, avoid the obstacle compared to other methods (\textit{w/o CBF}). In fact, in this scenario, learning occurs in two phases.

\textit{Goal-reaching phase}: The value functions of the trained agent at $10^6$ training steps and the CBF Filter policy at $5 \cdot 10^6$ training steps in Figure \ref{fig:value_func} are almost identical, indicating that the CBF Decay agent first learns to reach the goal. 

\textit{Obstacle avoidance phase}: Toward the end of training, the agent also learns to avoid obstacles, in Figure \ref{fig:rew_ep_length} we can see that the activation percentage of CBF Decay approaches zero, highlighting the agent’s ability to navigate  without interventions. This is also shown by the negative values in the obstacle region of the value function at $5 \cdot 10^6$ training steps in Figure \ref{fig:value_func} indicating that it learned to avoid obstacles.

These behaviors are visually confirmed in Figure \ref{fig:traj} (\textit{CBF Decay}), where trajectories demonstrate goal-reaching capabilities and deflection from obstacle. Additionally, we can observe in Figure \ref{fig:traj} that CBF Decay also learned to recover from collision with the obstacle, exit the unsafe region and eventually reach the goal. 


\textbf{Sim2Real.} Figure \ref{fig:traj} Demonstrate that simulation trajectories closely match real-world results, thus confirming that despite training under a CBF with a simplistic unicycle model, it was able to abstract the complexity of four wheel differential drive robot dynamics. 


The key takeaway from our results is that all methods learning under CBF intervention (both CBF Filter and CBF Decay) enable the agent to solve the task. Specifically, CBF Decay learns to recover from collisions without explicit recovery mechanisms, this is due to the fact that we allow it to be gradually aware of the catastrophic region.




\section{CONCLUSIONS}

CBF Decay approach enables learning to solve the task and  avoid collisions with the catastrophic zones, similar to training wheels being gradually removed when children are learning to ride bicycle. It appears that tasks containing reward designs based on continuous negative values (Both designed in SAC and CBF Reward) pose a challenge for reward-based RL, indicating that scalar rewards might not be sufficient \cite{rewIsEnough}.
In essence, we wish for the policy to favor certain trajectories over others and we believe that preference-based reinforcement learning could be a key direction for further research into learning to avoid these catastrophic regions.
As future work, we could provide a safety based privileged information (e.g; CBF Interventions) during training in simulation for the actor-critic framework, just like children are aware of the training wheels during learning. Additionally, we could develop a direct real-world learning pipeline with provable safety guarantees, building on our contribution of seamlessly integrating RL and CBF.







\section*{ACKNOWLEDGMENT}

The authors would like to thank Yann Bouteiller for their insights, Jana Pavlasek for their thoughtful feedback, and the entire MIST lab for their support.


\bibliography{root.bib}
\bibliographystyle{abbrv}

\end{document}